\documentclass[letterpaper, 10 pt, conference]{ieeeconf}  % Comment this line out if you need a4paper

\IEEEoverridecommandlockouts                              % This command is only needed if 
\usepackage{cite}
\usepackage{amsmath,amssymb,amsfonts}
\usepackage{algorithmic}
\usepackage{graphicx}
\usepackage{textcomp}
\usepackage{xcolor}
\usepackage{hyperref}
\usepackage{csquotes}
\usepackage{xcolor}
\usepackage{multirow}
\usepackage{tabularx}
\usepackage{listings}
\usepackage{xcolor}  
\usepackage{subfig}

\title{\LARGE \bf
Come Closer: The Effects of Robot Personality on Human Proxemics Behaviours
}

\author{Meriam Moujahid$^{1}$, David A. Robb, Christian Dondrup and Helen Hastie % <-this % stops a space
 \thanks{%*This work was funded and supported by the UKRI Node on Trust (EP/V026682/1).
 For the purpose of open access, the author has applied a Creative Commons Attribution (CC
BY) license to any Author Accepted Manuscript version arising.}% <-this % stops a space
 \thanks{$^{1}$School of Mathematical and Computer Sciences, Heriot-Watt University, EH14 4AS, UK.
 {\tt\small \{M.Moujahid, d.a.robb, c.dondrup, h.hastie\}@hw.ac.uk}}%
}

\begin{document}

\maketitle
\thispagestyle{empty}
\pagestyle{empty}

%%%%%%%%%%%%%%%%%%%%%%%%%%%%%%%%%%%%%%%%%%%%%%%%%%%%%%%%%%%%%%%%%%%%%%%%%%%%%%%%
\begin{abstract}
Social Robots in human environments need to be able to reason about their physical surroundings while interacting with people. Furthermore, human proxemics behaviours around robots can indicate how people perceive the robots and can inform robot personality and interaction design. Here, we introduce Charlie, a situated robot receptionist that can interact with people using verbal and non-verbal communication in a dynamic environment, where users might enter or leave the scene at any time. The robot receptionist is stationary and cannot navigate. Therefore, people have full control over their personal space as they are the ones approaching the robot. We investigated the influence of different apparent robot personalities on the proxemics behaviours of the humans.  
The results indicate that different types of robot personalities, specifically introversion and extroversion, can influence human proxemics behaviours. Participants maintained shorter distances with the introvert robot receptionist, compared to the extrovert robot. Interestingly, we observed that human-robot proxemics were not the same as typical human-human interpersonal distances, as defined in the literature. We therefore propose new proxemics zones for human-robot interaction. 

\end{abstract}
\vspace{-1mm}

%%%%%%%%%%%%%%%%%%%%%%%%%%%%%%%%%%%%%%%%%%%%%%%%%%%%%%%%%%%%%%%%%%%%%%%%%%%%%%%%
\section{INTRODUCTION}
\vspace{-1mm}
Social robots are entering our public and social spaces, where they need to be human-aware and follow social etiquette. Social robots can be valuable in the service industry, such as  receptionists \cite{moujahid2022demonstration}, baristas \cite{lim2022demonstration}, and bartenders \cite{keizer2014machine}.
If we observe how people interact with each other in physically situated face-to-face settings, the interaction goes well beyond the spoken words. Hence, the physical space in which the interaction takes place can have implications for the design of effective human-robot interaction. For such an effective interaction with humans, robots need to perceive and reason deeply about their physical surroundings, understand the physics of human interaction and engage in fluid interaction with humans in their physical environment \cite{andrist2020accelerating}. One of the commonly used measures to understand human social behaviours in a physical space is distance and positioning between interaction partners, often referred to as \emph{proxemics}.
\vspace{-1mm}

Initial work on the concept of proxemics \cite{hall1959silent} provided a systemic basis for research into social and personal spaces between humans. It was demonstrated that social spaces substantially reflect and influence social relationships and the attitudes of
people towards each other. It is not unreasonable to assume that such proxemics behaviours
might also hold true in human-robot interaction (HRI). Furthermore, it is essential to understand how design decisions for social robots could influence human proxemics behaviours around these robots.  To this end, this work looks at evaluating the influence of different robot personalities, i.e. introvert vs extrovert, on proxemics using a fixed position robot. Furthermore, we compare these proxemics to human-human interpersonal distances.

According to E. Howarth \cite{doi:10.2466/pr0.1969.24.2.415}, it is preferred to have extrovert traits when working in service industry jobs, this is valid for the receptionist role.
Previous work \cite{neff2010evaluating} 
demonstrated that manipulating linguistic and prosodic cues of a social robot can be used to portray extroversion and introversion personalities traits for the robot. Furthermore, an extrovert robot has been shown to be preferred and trusted over an introvert one \cite{lim2022we}. However, these studies did not delve into the proxemics aspect.

\begin{figure}[htbp]
\centerline{\includegraphics[scale=0.1]{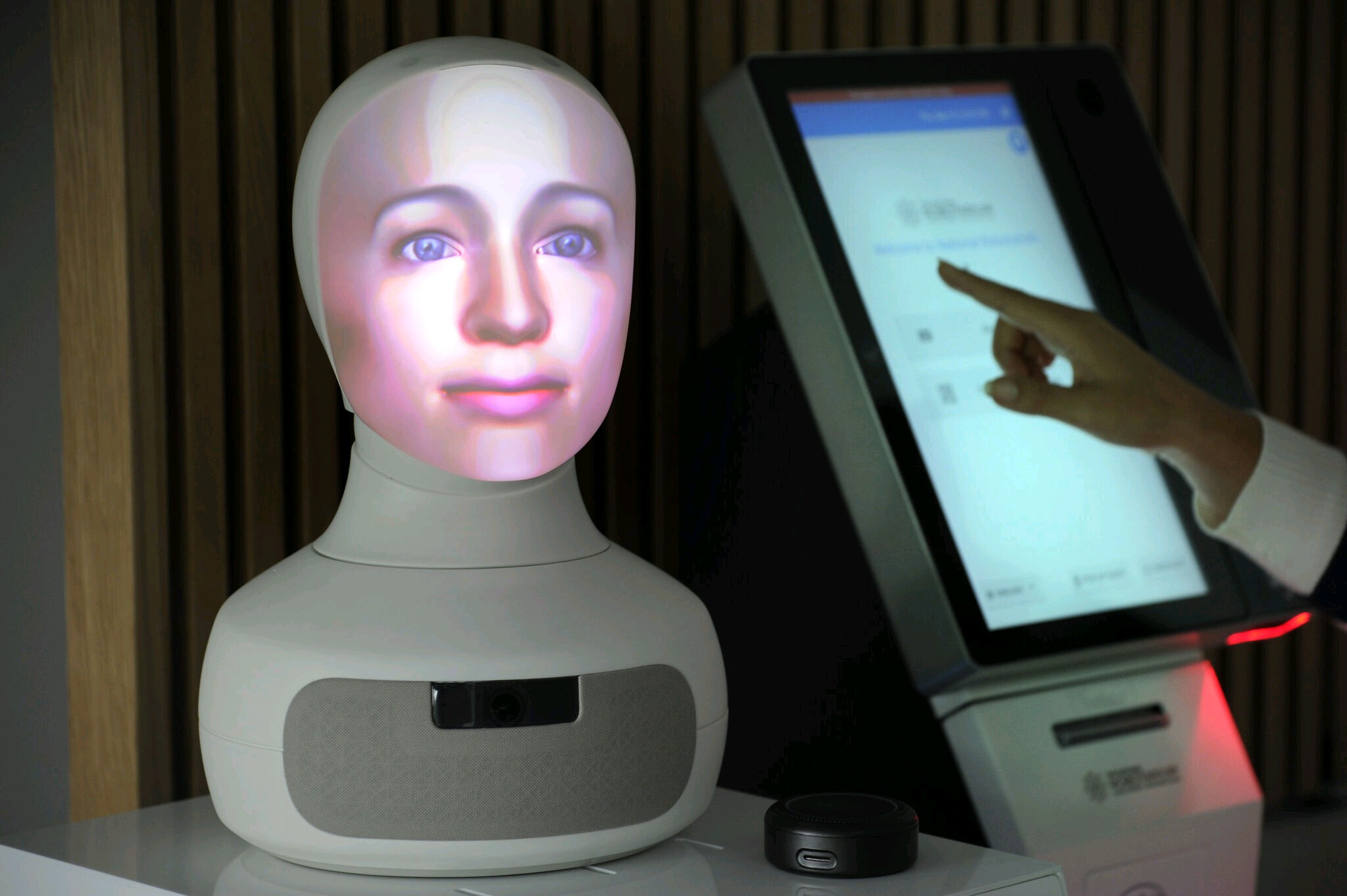}}
\caption{Furhat Robot Receptionist connected to a visitor management system and a TDS kiosk for visitors' check-in and check-out.}
\label{fig:receptionist}
\end{figure}

Our contributions in this paper include a unique study \enquote{in the wild} showing that human-robot proxemics are influenced by a robot's displayed personality. We use a Furhat\footnote{https://furhatrobotics.com} robot receptionist, connected to a visitor management system and a TDS kiosk\footnote{https://www.timedatasecurity.com/product/tds-visitor}, which is a self-service visitor check-in kiosk, as shown in Figure\ref{fig:receptionist}. The robot is able to reason about its physical surroundings and interact using a combination of verbal and non-verbal communication \cite{moujahid2022multi}.
We address two broad research questions (RQ):

RQ1: Can the robot's personality influence human proxemics behaviours?

RQ2: Are Human-robot interpersonal distances different to those found for human-human interpersonal distances?

\pagebreak

\section{BACKGROUND}

Edward T. Hall \cite{hall1966hidden} introduced the theory of proxemics, which refers to the personal space that people maintain around themselves. This space differs depending on the social setting and their cultural backgrounds. Michael Argyle \cite{argyle2013bodily} suggested the intimacy equilibrium model, which links mutual gaze and proxemics behaviours \cite{argyle1965eye}. This model illustrates how people react to the violation of their personal space by reducing mutual gaze and/or moving backward. 

A person's space is not only a physical buffer zone, but also a
psychological one \cite{dosey1969personal}. The invasion of our personal space can be uncomfortable and disquieting. Such a breach occurs when our intimate zone (0.00m to 0.45m) is violated by someone who is not an intimate connection. The ideal zone reserved for personal interactions among good friends or family is between 0.45 to 1.2, while the social zone reserved for interactions among acquaintances is between 1.2m and 2.1m \cite{hall1966hidden}. These guidelines used in social psychology are based on human-human interaction, and can only map to human-robot interaction if we concede that a robot is perceived as a social actor and comparable to a human \cite{nass1994computers} \cite{reeves1996people}. However, people do not always interact with robots in the same way that they interact with each other \cite{mumm2011human} \cite{sardar2012don}. Previous research suggests that humans approach robots with distances reserved for a close acquaintance or a family member \cite{walters2008human}, which is considered within the intimate zone. 

In human-robot interactions, there are many factors that could influence human proxemics behaviours. Some factors are related to the robots such as its voice \cite{walters2008human}, appearance  \cite{syrdal2008sharing}, speed \cite{butler2001psychological}, and height \cite{walters2008design}. Other aspects are related to humans such as their age \cite{childrenproxemics}, personalities \cite{syrdal2006doing} \cite{walters2005influence}, prior experience with a robot \cite{walters2008design}, gender \cite{syrdal2007personalized} or social norms \cite{argyle2013bodily}. Thus, it is critical to comprehend which robot design decisions impact human proxemics behaviours around robots, in order to enhance interactions between robots and humans.

It is yet unclear if robot personality can have any effects on human proxemics behaviours, given that it has been shown that we can model certain personality traits \cite{lim2022we}.
In this study, we test if proxemics vary depending on the robot personality type. Previous research on robot personality aims attention at the facet of extroversion, for being one of the easier personality traits to exhibit, even in shorter interactions \cite{robotpersona123}. On that account, we deemed it feasible to focus on extroversion and introversion as types of robot personalities.

Commonly, introverts are monotone and use fewer pauses and hesitations \cite{furnham1990handbook}, while extroverts use extensive vocabulary and speak with a faster speech rate and louder voice, in addition to higher fundamental frequency and broader frequency range \cite{pittam1994voice}, \cite{tusing2000sounds}. 
Earlier work \cite{modellingpersonality} \cite{personalitymatching} demonstrates how calibrating prosody can portray extroversion and introversion as personality traits in artificial agents, and can influence how we perceive their personality.

Other previous research \cite{pitchreceptionist} explored the influence of voice pitch on how people perceived a social robot receptionist. The findings point out that a
high-pitched robot with a female voice was rated
more attractive, emotional, and extrovert. 

Similarly, the voice of the robot can influence
approaching distances \cite{walters2008human}. Trovato et al. \cite{trovato2018sound} demonstrate that any uncomfortable noise produced by a robot can lead to greater human-robot distance when people approach it. However, this proxemics effect can be masked by adding ambient background music.

According to Bhagya et al. \cite{bhagya2019exploratory}, proxemics distances preferred by humans increase with the increased volume of the internal noises (i.e.: machine noise) of a robot. However, this was not tested with a robot that uses advanced speech capabilities and natural language. Macmillan \cite{macmillan199074} and  Chepesiuk  \cite{chepesiuk2005decibel} demonstrate that the ideal sound of a normal conversation is about 60dB, while any sound greater than 70db will be considered annoying noise for human ears, which might result in changing proxemics behaviours.
During this experiment, we therefore kept the volume of the robot speaker between 58dB and 64dB, measured using a sound meter.

\section{METHODOLOGY}

As mentioned above, Lim et al. \cite{lim2022we} showed that one can successfully
represent extroversion and introversion using linguistic features and vocal cues, they then went on to show that an extrovert robot is preferred and trusted more than an introvert robot in a robot barista setting. We followed a similar approach to Lim et al. \cite{lim2022we}, using prosodic parameters: volume, pitch, and tempo. We created two robot personalities with different traits of extroversion and introversion. However, as the context and domain of a robot receptionist was different to the robot barista, in their study, we relied on other previous work \cite{tusing2000sounds} to add pragmatic \cite{neff2010evaluating} and lexical/syntactic \cite{mairesse2007personage} differences between the two personalities. 

We explore how different types of robot personalities, specifically introversion and extroversion, can lead to different human
proxemics behaviours. Furthermore, in the wild studies can give us new insights into spontaneous interactions with robots, giving people the freedom to approach the robot (or not) at a comfortable distance.
This work addresses these questions and fills this gap by comparing how an introvert vs. extrovert robot receptionist will affect proxemics behaviours. 

\section{EXPERIMENTAL DESIGN AND SETUP}

\begin{figure}[htbp]
\centerline{\includegraphics[scale=0.45]{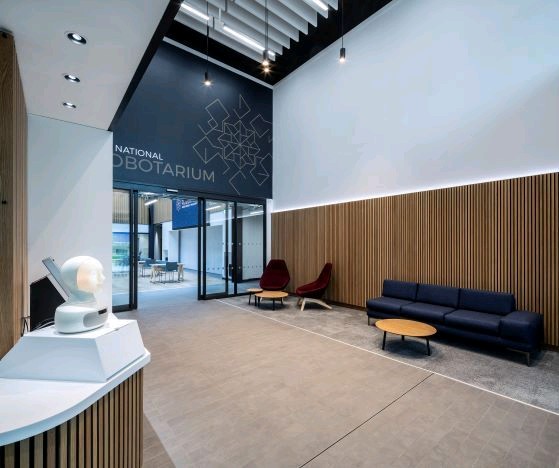}}
\caption{The reception area in the National Robotarium where the robot receptionist is placed}
\label{fig:receptionphoto}
\end{figure}

Charlie, the Robot Receptionist set-up, consists of a Furhat robot \cite{Furhat12} from Furhat Robotics\footnote{https://furhatrobotics.com} that is connected to a visitor management system through an API, in order to get information about calendar, meetings, employee profile and to guide visitors with the check-in and check-out process.
The robot is placed in the reception area of the UK National Robotarium\footnote{https://thenationalrobotarium.com}, as shown on Figure \ref{fig:receptionphoto}. The robot can interact with users using verbal and non-verbal communication and has advanced natural language understanding and dialogue management capabilities. The receptionist can help visitors with the registration process in order to print a badge, and give them information about directions, news and events.
The Robot Receptionist is aware of its physical surroundings. Furthermore, it can reason and react to events in the physical space around itself. %%For further details about the robot set-up, see [ANON]. 

\subsection{Conditions}

We created two robot personalities with different traits of extroversion and introversion, by manipulating the lexicon (See Table \ref{tab:table1}), volume, pitch and
tempo (following \cite{lim2022we}) as follows:
\begin{enumerate}
 \item The Introvert: The robot speech was given a lower pitch (-20\%), volume (-6dB) and speaking rate (-20\%).
 \item The Extrovert: The robot speech was given a higher pitch (+20\%), volume (+6dB) and speaking rate (+20\%).
 
\end{enumerate}

\begin{table}
% \scalebox{0.5}{
{\small % 
\caption{\label{tab:table1} Extrovert and Introvert statements and linguistic features applied to the robot interaction}
\centering
\scalebox{0.75}{
\begin{tabularx}{0.635\textwidth} {| p{4cm} | p{4cm} | p{2cm} |}

 \hline
 \textbf{Introvert Robot Receptionist} & \textbf{Extrovert Robot Receptionist} & \textbf{Linguistic feature}\\ % & \textbf{N}\\
 \hline

    Hello! Welcome to the National Robotarium! I am Charlie the receptionist! I will be happy to assist you while entering your details.   & Hi! Welcome to the National Robotarium! \textbf{Lovely to see you} ! \textbf{By the way}, I am Charlie the receptionist. I will be happy to assist you while entering your details! & Subject implicitness, positive emotion, self-reference. \\ 
\hline
    Just remember to speak when my green light is on! That's when I can hear you! & \textbf{Actually, It will be great if} you can speak when my green light is on! That's when I can hear you! & Positive emotion, acknowledgement.\\
\hline 
    Did you receive an invitation email with a QR code prior to your visit?  & \textbf{Firstly, I would like to know if} you received an invitation email with a QR code prior to your visit? & Self-reference \\
\hline

Can you please select the option, check in, on the kiosk, then scan the QR code? & \textbf{That's Great! Basically, I just need you to} select the option, check in, on the kiosk, then scan the QR code?  & Positive emotion, acknowledgement, self-reference\\
\hline
    You will need To register! Please press the first button that says Register! & \textbf{No worries at all ! Actually}, I just need you to register on the kiosk ! Please press the first button that says register. & Positive emotion, acknowledgement, self-reference.\\
    
\hline
    Sorry! I didn't hear you. Would you speak a little louder when my green light is on please? & Sorry I didn't \textbf{catch that} ! Is it possible for you to raise your voice, a little bit please, When my green light is on? Just so \textbf{ I can hear you clearly} ! & informal, self-reference, acknowledgement \\

\hline
    Bye for now. & \textbf{Have a nice day}! Bye! & Positive emotion \\
    
\hline

\end{tabularx}}

}
\end{table}

We observed 3 types of interaction with the robot receptionist, therefore we classified participants into 3 groups:
\begin{enumerate}
 \item \textbf{Verbal interactions}: People used verbal communication to interact with the robot.
 \item \textbf{Non-verbal interactions}: People were standing in the robot field of view, and looking towards the robot for more than 15 seconds but without talking.
 \item \textbf{No interaction}: People just pass by, without showing any interest. These were identified by a very short time period (less than 15 seconds) between being detected in the scene and leaving the scene.
\end{enumerate}
The experiment was between-participants by design. The independent variables are thus:

- The \textbf{Robot Personality} as a manipulated independent variable with 2 levels: Introversion vs Extroversion policy. 

- The \textbf{Type of Interaction} as a non manipulated independent variable with 3 levels: Verbal, Non-verbal or No interaction

The main dependent variable for the experiment is the \textbf{Shortest Distance} between the robot and the user during the entire interaction.

During the experiment, a total of 120 participants took part (60 interactions for each personality) over 6 weeks. All interaction data were collected with the exact user position coordinates and orientation at 5 different time stamps with {\textit{$t_1$} when the user is first detected and {\textit{$t_5$} when the user leaves the robot's scene. We also collected the robot-transcribed utterances, and the duration of the interaction. The participants' visual or personal data were not collected. No audio or video data was collected, also no demographic information was collected as the experiment was in the wild. The procedure was ethically approved by our institution's ethics board.

\subsection{Hypotheses}

In order to address our research questions, we formulated the following hypotheses:

\begin{itemize}
    \item \textbf{H1:} There is a significant difference for the shortest distance between the two robot personality types (Introvert/Extrovert)

    \item \textbf{H2:} There is significant difference for the shortest distance between the types of interaction (Verbal, Non-verbal and No Interaction)

    \item \textbf{H3:} There is a significant interaction effect between robot personality types and the types of interaction, with respect to shortest distance 

\end{itemize}
\clearpage
\pagebreak

\section{RESULTS AND ANALYSIS} % present results in order H1-H2-H3

Firstly, we grouped the participants into 3 groups based on the Type of Interaction they had with the robot.
Table \ref{tab:table2} represents the number and percentage of users engaging in each Type of Interaction.

\begin{table}
% \scalebox{0.5}{
{\small % 
\caption{\label{tab:table2}The number of users and types of their interactions with the Robot Receptionist using different policies}
\centering
\scalebox{0.75}{
\begin{tabularx}{0.64\textwidth} {| p{3cm} | p{1cm} | p{1,70cm} | p{1cm} | p{1,70cm} |}

 \hline
 \textbf{Type of Interactions} & \multicolumn{2}{|c|}{\textbf{With the introvert Robot}} & \multicolumn{2}{|c|}{\textbf{With the extrovert Robot}}\\ % & \textbf{N}\\
 \hline
  
       & \textbf{Number} & \textbf{Percentage} & \textbf{Number} & \textbf{Percentage} \\ 
\hline
    Participants using Verbal Interaction  & 22 & 36.66\% & 18 & 30.00\% \\ 
\hline
    Participants using Non-verbal Interaction  & 19 & 31.66\% & 20 & 33.33\% \\
\hline
    No Interaction  & 19 & 31.66\% & 22 & 36.66\% \\
\hline
\end{tabularx}}
}
\end{table}

\begin{figure}[htbp]
\centerline{\includegraphics[scale=0.21]{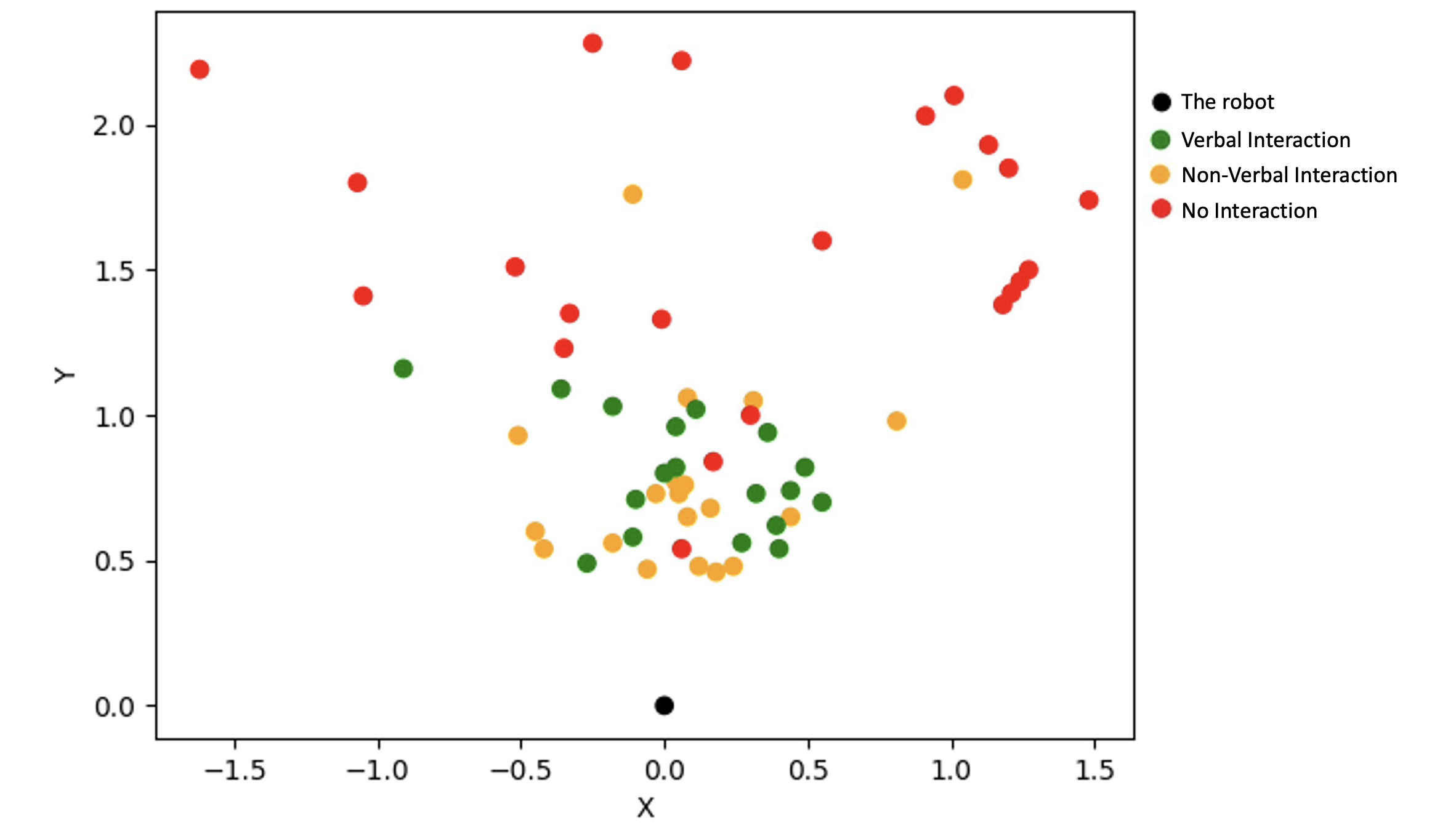}}
\caption{Extrovert Robot: Scatter plot of the users X and Y coordinates showing the users' position on the reception scene in front of the Extrovert Robot receptionist. (The black dot represents the robot's position at (0.0,0.0))}
\label{fig:userinteraction}

\centerline{\includegraphics[scale=0.21]{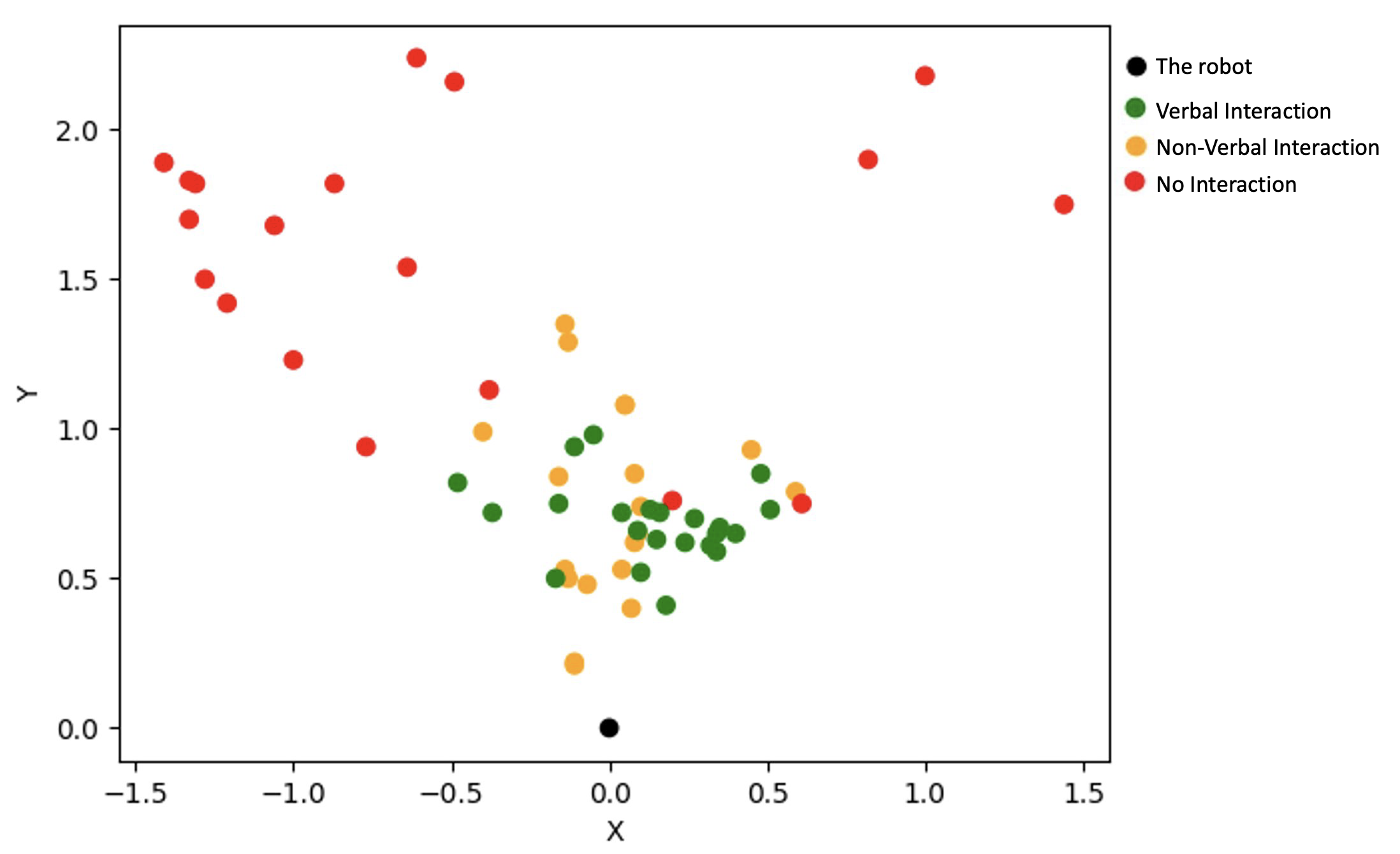}}
\caption{Introvert Robot: Scatter plot of the users X and Y coordinates showing the users' position on the reception scene in front of the Introvert Robot receptionist. (The black dot represents the robot's position at (0.0,0.0))}
\label{fig:userinteraction2}
\end{figure}

Using the users' location coordinates, we computed the Euclidean distance for each logged interaction.
We used the Shortest Distance between the participant and the robot. The Shortest Distance is preferred over average distance as it is more accurate \cite{bailenson2001equilibrium}. Furthermore, the users used the kiosk next to the robot (Figure\ref{fig:receptionist}), which will result in them standing at a specific distance from the robot for an unequal amount of time. This would lead to inconsistent results when calculating the average distance.
Figures \ref{fig:userinteraction} and \ref{fig:userinteraction2} show plots of users interacting with each personality.
%DR 
%In order to address our hypothesis, we calculated the means of the Shortest Distance in all conditions, then in each condition, 
Table \ref{tab:table3} shows the descriptive statistics of the Shortest Distance in each of the two personality conditions across all three Types of Interaction and over all the interactions as a whole. 
%DR deleting this
%After a normality test, we did a two-way ANOVA  then post hoc analysis followed by an independent samples t-test.
\begin{table}
% \scalebox{0.5}{
{\small % 
\caption{\label{tab:table3} Shortest Distance maintained by participants while interacting with the Robot Receptionist using different personalities}
\centering
\scalebox{0.7}{
\begin{tabularx}{0.685\textwidth} {| p{1.8cm} | p{0.8cm} | p{0.8cm} | p{0.8cm} | p{0.8cm} | p{0.8cm} | p{0.8cm} | p{0.8cm} | p{0.8cm}|}

 \hline
 \textbf{Type of Interactions} & \multicolumn{4}{|c|}{\textbf{Introvert Robot}} & \multicolumn{4}{|c|}{\textbf{Extrovert Robot}}\\ % & \textbf{N}\\
 \hline
  
       & \textbf{Min} & \textbf{Max} & \textbf{Mean} & \textbf{Std Dev} & \textbf{Min} & \textbf{Max} &\textbf{Mean}  & \textbf{Std Dev}\\ 
\hline
    Participants using Verbal Interaction  & 0.41m & 0.91m & \textbf{0.61m} & 0.12 & 0.55m & 1.47m & \textbf{1.19m} & 0.46\\ 
\hline
    Participants using Non-verbal Interaction & 0.23m & 1.35m & 0.64m & 0.32 & 0.47m & 2.08m & 1.01m & 0.51\\
\hline 
    No Interaction  & 0.78m & 2.34m & 1.78m & 0.48 & 0.54m & 2.72m & 1.79m & 0.54\\
\hline 
\hline 
    All Interactions  & 0.23m & 2.34m & 0.99m & 0.63 & 0.47m & 2.72m & 1.36m & 0.60\\
\hline

\end{tabularx}}

}
\end{table}

 A factorial ANOVA  (also known as two-way ANOVA) was carried out in IBM SPSS version 28.  
 
\textit{Robot Personality:}
There was a significant main effect of Robot Personality, with participants’ Shortest Distance being closer to the Introvert Robot (Mean, 0.99) than the Extrovert Robot (Mean, 1.36), \textit{F}(1,114) = 16.76, \textit{p}\textless.001. 
We can therefore accept H1.
Furthermore, the ANOVA showed that the Shortest distance for the Extrovert group was significantly higher than the Introvert group as shown on Figure \ref{fig:comparison}.

 \begin{figure}[htbp]
\centerline{\includegraphics[scale=0.3]{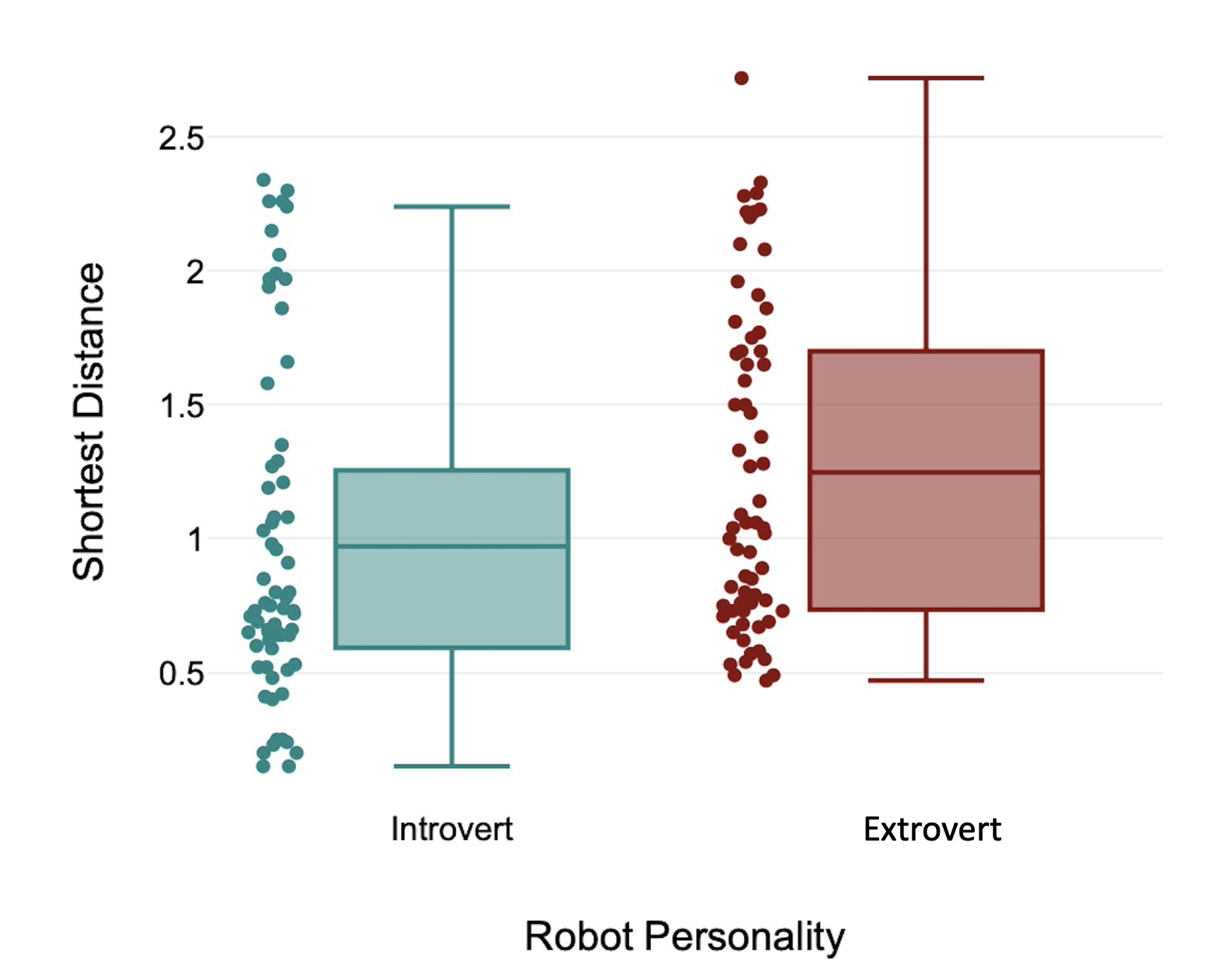}}
%\centerline{\includegraphics[scale=0.3]{results/t-test.png}}
%\caption{One-tailed t-test results showing that the users approached the introvert robot within a significantly shorter distance compared to the extrovert robot, the solid horizontal lines mark the mean and the dotted lines mark the medians}
\caption{Box plots showing that the users approached the introvert robot within a significantly shorter distance compared to the extrovert robot. (boxes show the means and the two middle quartiles, while the whiskers show maximums and minimums other than any notional outliers)}
\label{fig:comparison}
\end{figure}
\newpage
\textit{Type of Interaction:}
There was a significant main effect of Type of Interaction on the participants’ Shortest Distance from the Robot, \textit{F}(2,114) = 61.69, \textit{p}\textless.001. 
We can therefore accept H2.
Post hoc tests using Bonferroni’s correction for multiple comparisons, revealed that the participants’ Shortest Distance was significantly shorter for those in the Verbal (Mean, 0.88) and Non-verbal (Mean, 0.84) Interaction groups (\textit{p}\textless0.001 in both cases) compared to those in the No Interaction group (Mean, 1.79).  There was no significant difference in Shortest Distance between participants in the Verbal and Non-Verbal Interaction groups (\textit{p} =.902). 

\textit{Interaction Effect:}
There was a significant interaction effect between the Robot Personality and the Type of Interaction measured by Shortest Distance \textit{F}(2,114) = 4.57, \textit{p} =.012. This effect indicates that participants' Shortest Distance from the Introvert Robot and the Extrovert Robot was affected differently by their Type of Interaction.  
We can therefore accept H3.
%The chart in 
Figure \ref{fig:Anova} shows the Means of the Shortest Distance for each group along with their 95\% confidence intervals. We can see this indicates that the Shortest Distance for those participants in both Verbal and Non-Verbal groups was influenced by Robot personality (with them being closer to the Introvert Robot). On the other hand, the Shortest Distance of those who did not interact with the robot was not affected by robot personality. 

\begin{figure}[htbp]
\centerline{\includegraphics[scale=0.875]{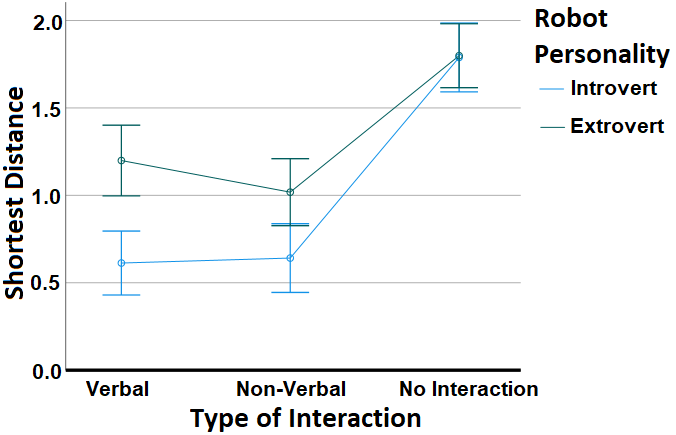}}
%\centerline{\includegraphics[scale=0.3]{results/t-test.png}}
%\caption{One-tailed t-test results showing that the users approached the introvert robot within a significantly shorter distance compared to the extrovert robot, the solid horizontal lines mark the mean and the dotted lines mark the medians}
\caption{Mean Shortest Distances (in metres) for the different groups, along with the 95\% confidence intervals.}
\label{fig:Anova}
\end{figure}

\section{Discussion}
\textit{RQ1: Can the robot's personality influence human proxemics behaviours?}

In support of H1, %we can validate that different types of robot personalities,
the results show that different types of robot personalities, in this case introversion and extroversion, can lead to different human proxemics behaviours.
Furthermore, the distance was shorter with the introvert robot, which %convey 
indicates that participants were more comfortable standing closer to the introvert robot.
Shorter proxemics could be linked to positive feelings about the robots.

There are wider implications of
these results for %the 
human-robot interaction theory and the design of a social robot and specifically for designing its personality.

\textit{RQ2: Are Human-robot interpersonal distances different to those found for human-human interpersonal distances?}

By accepting H2, we demonstrate that the Shortest Distance between human and robot was different to what would be expected between two humans, based on the Type of Interactions %and 
regardless of the robot personality.
We notice that none of the distance ranges fall within the human-human social zone, which indicates that people did not follow the same human-human proxemics rules \cite{hall1966hidden} to interact with the robot, as shown on Figure \ref{fig:proxemicshri}.

\begin{figure}[htbp]
\centerline{\includegraphics[scale=0.4]{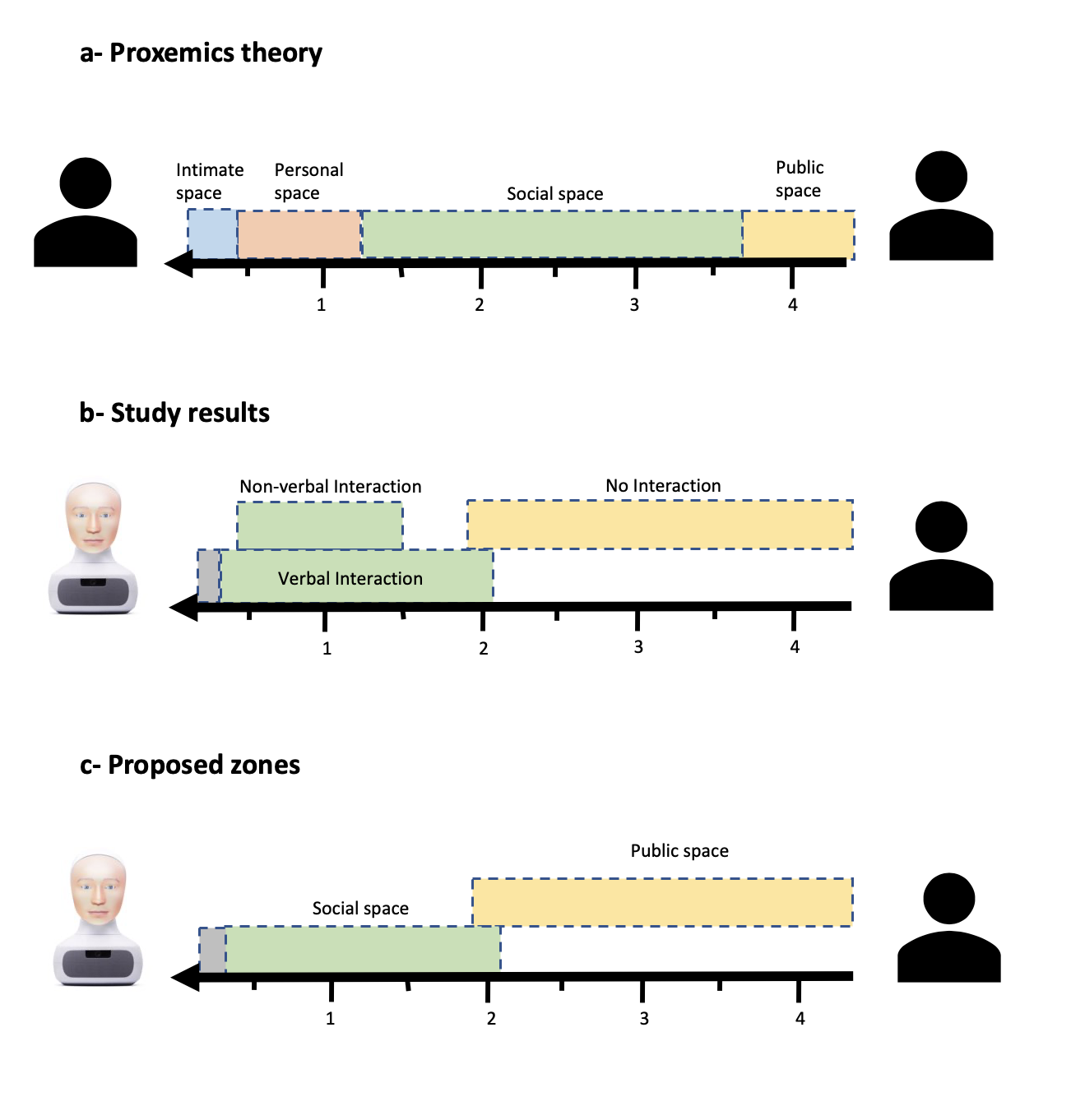}}
\caption{Human-human interaction zones according to proxemics theory compared to interaction zones with the Robot Receptionist}
\label{fig:proxemicshri}
\end{figure}

Participants using verbal communication kept a distance between 0.41m and 1.47m. Nonetheless, this distance does not fit with the social zone for human-human interactions (1.2m to 3.6m). 
It is however very close to the personal zone (0.45m to 1.2m) as considered by Hall's proxemics theory \cite{hall1966hidden}.

Similarly, participants using non-verbal communication kept a distance between 0.23m and 2.08m. This distance overlap between the social zone for human-human interactions (1.2m to 3.6m) and the personal zone (0.45m to 1.2m).

Lastly, participants who avoided interacting with the robot used a zone in the reception area, which is $ \ge  $0.78m away from the robot. This is again considered in human-human interaction as being in a personal and social zone.

These results contradict the media equation theory \cite{nass1994computers} \cite{reeves1996media} and Hall's proxemics theory \cite{hall1966hidden}. While these theories rely on principles from social psychology and social science, our results suggest that these theories do not always map to human robot-interaction. Therefore, in Figure \ref{fig:proxemicshri}c, we propose a new set of proxemics zones for this form of interaction with a stationary robot.

Our findings suggest the comfortable distance people like to keep when approaching the robot might vary between 0.23m and 2.08m (\ref{tab:table3}, second row, Non-verbal Interaction, Introvert Robot, Min, 0.23m and Extrovert Robot, Max, 2.08m).

%DR addin this see what you think. Not sure if this is the exact right place for it
The grey coloured zones in Figure \ref{fig:proxemicshri} b and c, represent the narrow zone very close to the robot, for which we have no data. The vision system within the Robot itself is unable to sense and gather accurate data within this zone very close to it. Thus, it represents what might be termed a blind spot. Thus we have no data on whether or not people were approaching closer than this. For this reason, we represent this space in grey as unknown. This close zone is a space for future research.

As this experiment was done \enquote{in the wild}, participants were interacting with the robot in a spontaneous way.
Some participants were using only non-verbal cues. While the robot was verbally interactive, they chose not to verbally respond to the robot. This behaviour could be explained by people meeting the robot for the first time, or being curious about the robot and observing its behaviour. 

This type of behaviour will be unusual in the context of human-human interaction, again contradicting the media equation theory. However, this suggests that the non-verbal social communication of the robot could be improved to also respond using more non-verbal cues in these types of situations.

\section{CONCLUSION AND FUTURE WORK}

We show that the distance between robot and human can be different
depending on the robot personality. To demonstrate this, we focused on introversion and extroversion as personality traits. We provide empirical evidence that people who had a verbal interaction with the robot maintained a shorter distance with an Introvert robot compared to the Extrovert version. 

Overall, the distance people maintained with the robot, regardless of the robot personality, was between 0.41m and 1.47m for verbal interaction, and between 0.23m to 2.08m for non-verbal interaction.  This range is overlapping between human-human social zone and personal zone. These results imply that proxemics theory does not %scale 
map directly
to human-robot interactions as people do not interact with robots in the same way
they interact with each other.

There are broader implications of these results in the design of social robots and interactions. For instance, factors related to the robot personality traits can influence human proxemics behaviours around these robots.

In the future, further data will be collected to analyse other parameters that can influence proxemics behaviours, such as the robot's physical aspects and gestures. Moreover, we want to analyse further variables such as the duration of the interaction and look into the human trajectories in more detail, and how these proxemics behaviours change over time. 

%\section*{Acknowledgements}
%This work was funded and supported by the UKRI Node on Trust (EP/V026682/1) \url{https://trust.tas.ac.uk}.
%This work was also supported by the European Commission under the Horizon 2020 framework program for Research and Innovation (H2020-ICT-2019-2, GA \#871245), SPRING project. \url{https://spring-h2020.eu}. 
%\clearpage
\pagebreak
\section*{Acknowledgements}
This work was funded and supported by the UKRI Node on Trust (EP/V026682/1) \url{https://trust.tas.ac.uk}.
We would like to thank the National Robotarium Engineers \url{https://thenationalrobotarium.com} and the TDS support team \url{https://www.timedatasecurity.com/product/tds-visitor}.

\bibliographystyle{IEEEtran}

\bibliography{root}

\begin{thebibliography}{10}
\providecommand{\url}[1]{#1}
\csname url@samestyle\endcsname
\providecommand{\newblock}{\relax}
\providecommand{\bibinfo}[2]{#2}
\providecommand{\BIBentrySTDinterwordspacing}{\spaceskip=0pt\relax}
\providecommand{\BIBentryALTinterwordstretchfactor}{4}
\providecommand{\BIBentryALTinterwordspacing}{\spaceskip=\fontdimen2\font plus
\BIBentryALTinterwordstretchfactor\fontdimen3\font minus
  \fontdimen4\font\relax}
\providecommand{\BIBforeignlanguage}[2]{{%
\expandafter\ifx\csname l@#1\endcsname\relax
\typeout{** WARNING: IEEEtran.bst: No hyphenation pattern has been}%
\typeout{** loaded for the language `#1'. Using the pattern for}%
\typeout{** the default language instead.}%
\else
\language=\csname l@#1\endcsname
\fi
#2}}
\providecommand{\BIBdecl}{\relax}
\BIBdecl

\bibitem{moujahid2022demonstration}
M.~Moujahid, B.~Wilson, H.~Hastie, and O.~Lemon, ``Demonstration of a robot
  receptionist with multi-party situated interaction,'' in \emph{Proceedings of
  the 2022 17th ACM/IEEE International Conference on Human-Robot Interaction
  (HRI)}.\hskip 1em plus 0.5em minus 0.4em\relax IEEE, 2022, pp. 1202--1203.

\bibitem{lim2022demonstration}
M.~Y. Lim, J.~D.~A. Lopes, D.~A. Robb, B.~W. Wilson, M.~Moujahid, and
  H.~Hastie, ``Demonstration of a robo-barista for in the wild interactions,''
  in \emph{Proceedings of the 2022 17th ACM/IEEE International Conference on
  Human-Robot Interaction (HRI)}.\hskip 1em plus 0.5em minus 0.4em\relax IEEE,
  2022, pp. 1200--1201.

\bibitem{keizer2014machine}
S.~Keizer, M.~Ellen~Foster, Z.~Wang, and O.~Lemon, ``Machine learning for
  social multiparty human--robot interaction,'' \emph{Proceedings of the ACM
  transactions on interactive intelligent systems (TIIS)}, vol.~4, no.~3, pp.
  1--32, 2014.

\bibitem{andrist2020accelerating}
S.~Andrist and D.~Bohus, ``Accelerating the development of multimodal,
  integrative-ai systems with platform for situated intelligence,'' \emph{arXiv
  preprint arXiv:2010.06084}, 2020.

\bibitem{hall1959silent}
E.~T. Hall and T.~Hall, \emph{The silent language}.\hskip 1em plus 0.5em minus
  0.4em\relax Anchor books, 1959, vol. 948.

\bibitem{doi:10.2466/pr0.1969.24.2.415}
\BIBentryALTinterwordspacing
E.~Howarth, ``Expectations concerning occupations in relation to
  extraversion-introversion,'' \emph{Psychological Reports}, vol.~24, no.~2,
  pp. 415--418, 1969, pMID: 5809034. [Online]. Available:
  \url{https://doi.org/10.2466/pr0.1969.24.2.415}
\BIBentrySTDinterwordspacing

\bibitem{neff2010evaluating}
M.~Neff, Y.~Wang, R.~Abbott, and M.~Walker, ``Evaluating the effect of gesture
  and language on personality perception in conversational agents,'' in
  \emph{Proceedings of the International Conference on Intelligent Virtual
  Agents}.\hskip 1em plus 0.5em minus 0.4em\relax Springer, 2010, pp. 222--235.

\bibitem{lim2022we}
M.~Y. Lim, J.~D.~A. Lopes, D.~A. Robb, B.~W. Wilson, M.~Moujahid,
  E.~De~Pellegrin, and H.~Hastie, ``We are all individuals: The role of robot
  personality and human traits in trustworthy interaction,'' in
  \emph{Proceedings of the 2022 31st IEEE International Conference on Robot and
  Human Interactive Communication (RO-MAN)}.\hskip 1em plus 0.5em minus
  0.4em\relax IEEE, 2022, pp. 538--545.

\bibitem{moujahid2022multi}
M.~Moujahid, H.~Hastie, and O.~Lemon, ``Multi-party interaction with a robot
  receptionist,'' in \emph{Proceedings of the 2022 17th ACM/IEEE International
  Conference on Human-Robot Interaction (HRI)}.\hskip 1em plus 0.5em minus
  0.4em\relax IEEE, 2022, pp. 927--931.

\bibitem{hall1966hidden}
E.~T. Hall and E.~T. Hall, \emph{The hidden dimension}.\hskip 1em plus 0.5em
  minus 0.4em\relax Anchor, 1966, vol. 609.

\bibitem{argyle2013bodily}
M.~Argyle, \emph{Bodily communication}.\hskip 1em plus 0.5em minus 0.4em\relax
  Routledge, 2013.

\bibitem{argyle1965eye}
M.~Argyle and J.~Dean, ``Eye-contact, distance and affiliation,''
  \emph{Sociometry}, pp. 289--304, 1965.

\bibitem{dosey1969personal}
M.~A. Dosey and M.~Meisels, ``Personal space and self-protection.''
  \emph{Journal of personality and social psychology}, vol.~11, no.~2, p.~93,
  1969.

\bibitem{nass1994computers}
C.~Nass, J.~Steuer, and E.~R. Tauber, ``Computers are social actors,'' in
  \emph{Proceedings of the SIGCHI conference on Human factors in computing
  systems}, 1994, pp. 72--78.

\bibitem{reeves1996people}
B.~Reeves and C.~Nass, ``How people treat computers, television, and new media
  like real people and places,'' 1996.

\bibitem{mumm2011human}
J.~Mumm and B.~Mutlu, ``Human-robot proxemics: physical and psychological
  distancing in human-robot interaction,'' in \emph{Proceedings of the 6th
  international conference on Human-robot interaction}, 2011, pp. 331--338.

\bibitem{sardar2012don}
A.~Sardar, M.~Joosse, A.~Weiss, and V.~Evers, ``Don't stand so close to me:
  users' attitudinal and behavioral responses to personal space invasion by
  robots,'' in \emph{Proceedings of the seventh annual ACM/IEEE international
  conference on Human-Robot Interaction}, 2012, pp. 229--230.

\bibitem{walters2008human}
M.~L. Walters, D.~S. Syrdal, K.~L. Koay, K.~Dautenhahn, and R.~Te~Boekhorst,
  ``Human approach distances to a mechanical-looking robot with different robot
  voice styles,'' in \emph{Proceedings of the RO-MAN 2008-The 17th IEEE
  international symposium on robot and human interactive communication}.\hskip
  1em plus 0.5em minus 0.4em\relax IEEE, 2008, pp. 707--712.

\bibitem{syrdal2008sharing}
D.~S. Syrdal, K.~Dautenhahn, M.~L. Walters, and K.~L. Koay, ``Sharing spaces
  with robots in a home scenario-anthropomorphic attributions and their effect
  on proxemic expectations and evaluations in a live hri trial.'' in
  \emph{Proceedings of the AAAI fall symposium: AI in Eldercare: new solutions
  to old problems}, 2008, pp. 116--123.

\bibitem{butler2001psychological}
J.~T. Butler and A.~Agah, ``Psychological effects of behavior patterns of a
  mobile personal robot,'' \emph{Autonomous Robots}, vol.~10, no.~2, pp.
  185--202, 2001.

\bibitem{walters2008design}
M.~L. Walters, ``The design space for robot appearance and behaviour for social
  robot companions,'' Ph.D. dissertation, University of Hertfordshire, 2008.

\bibitem{childrenproxemics}
D.~Tokmurzina, N.~Sagitzhan, A.~Nurgaliyev, and A.~Sandygulova, ``Exploring
  child-robot proxemics,'' in \emph{Proceedings of the Companion of the 2018
  ACM/IEEE International Conference on Human-Robot Interaction}, 2018, pp.
  257--258.

\bibitem{syrdal2006doing}
D.~S. Syrdal, K.~Dautenhahn, S.~Woods, M.~L. Walters, and K.~L. Koay, ``'doing
  the right thing wrong'-personality and tolerance to uncomfortable robot
  approaches,'' in \emph{Proceedings of the ROMAN 2006-The 15th IEEE
  International Symposium on Robot and Human Interactive Communication}.\hskip
  1em plus 0.5em minus 0.4em\relax IEEE, 2006, pp. 183--188.

\bibitem{walters2005influence}
M.~L. Walters, K.~Dautenhahn, R.~Te~Boekhorst, K.~L. Koay, C.~Kaouri, S.~Woods,
  C.~Nehaniv, D.~Lee, and I.~Werry, ``The influence of subjects' personality
  traits on personal spatial zones in a human-robot interaction experiment,''
  in \emph{Proceedings of the ROMAN 2005. IEEE International Workshop on Robot
  and Human Interactive Communication, 2005.}\hskip 1em plus 0.5em minus
  0.4em\relax IEEE, 2005, pp. 347--352.

\bibitem{syrdal2007personalized}
D.~S. Syrdal, K.~L. Koay, M.~L. Walters, and K.~Dautenhahn, ``A personalized
  robot companion?-the role of individual differences on spatial preferences in
  hri scenarios,'' in \emph{Proceedings of the RO-MAN 2007-The 16th IEEE
  International Symposium on Robot and Human Interactive Communication}.\hskip
  1em plus 0.5em minus 0.4em\relax IEEE, 2007, pp. 1143--1148.

\bibitem{robotpersona123}
L.~Robert, ``Personality in the human robot interaction literature: A review
  and brief critique,'' in \emph{Robert, LP (2018). Personality in the Human
  Robot Interaction Literature: A Review and Brief Critique, Proceedings of the
  24th Americas Conference on Information Systems, Aug}, 2018, pp. 16--18.

\bibitem{furnham1990handbook}
A.~Furnham, ``Handbook of language and social psychology, chapter language and
  personality,'' 1990.

\bibitem{pittam1994voice}
J.~Pittam, \emph{Voice in social interaction}.\hskip 1em plus 0.5em minus
  0.4em\relax Sage, 1994, vol.~5.

\bibitem{tusing2000sounds}
K.~J. Tusing and J.~P. Dillard, ``The sounds of dominance. vocal precursors of
  perceived dominance during interpersonal influence,'' \emph{Human
  Communication Research}, vol.~26, no.~1, pp. 148--171, 2000.

\bibitem{modellingpersonality}
M.~Schmitz, A.~Kr{\"u}ger, and S.~Schmidt, ``Modelling personality in voices of
  talking products through prosodic parameters,'' in \emph{Proceedings of the
  12th international conference on Intelligent user interfaces}, 2007, pp.
  313--316.

\bibitem{personalitymatching}
A.~Tapus, C.~Ţăpuş, and M.~Matarić, ``User-robot personality matching and
  robot behavior adaptation for post-stroke rehabilitation therapy,''
  \emph{Intelligent Service Robotics}, vol.~1, pp. 169--183, 04 2008.

\bibitem{pitchreceptionist}
A.~Niculescu, B.~Van~Dijk, A.~Nijholt, and S.~L. See, ``The influence of voice
  pitch on the evaluation of a social robot receptionist,'' in
  \emph{Proceedings of the 2011 International Conference on User Science and
  Engineering (i-USEr)}.\hskip 1em plus 0.5em minus 0.4em\relax IEEE, 2011, pp.
  18--23.

\bibitem{trovato2018sound}
G.~Trovato, R.~Paredes, J.~Balvin, F.~Cuellar, N.~B. Thomsen, S.~Bech, and
  Z.-H. Tan, ``The sound or silence: investigating the influence of robot noise
  on proxemics,'' in \emph{Proceedings of the 2018 27th IEEE international
  symposium on robot and human interactive communication (RO-MAN)}.\hskip 1em
  plus 0.5em minus 0.4em\relax IEEE, 2018, pp. 713--718.

\bibitem{bhagya2019exploratory}
S.~Bhagya, P.~Samarakoon, M.~Viraj, J.~Muthugala, A.~Buddhika, P.~Jayasekara,
  and M.~R. Elara, ``An exploratory study on proxemics preferences of humans in
  accordance with attributes of service robots,'' in \emph{Proceedings of the
  2019 28th IEEE International Conference on Robot and Human Interactive
  Communication (RO-MAN)}.\hskip 1em plus 0.5em minus 0.4em\relax IEEE, 2019,
  pp. 1--7.

\bibitem{macmillan199074}
R.~Macmillan, ``74.20 decibel arithmetic,'' \emph{The Mathematical Gazette},
  vol.~74, no. 468, pp. 150--153, 1990.

\bibitem{chepesiuk2005decibel}
R.~Chepesiuk, ``Decibel hell: the effects of living in a noisy world,'' 2005.

\bibitem{mairesse2007personage}
F.~Mairesse and M.~Walker, ``Personage: Personality generation for dialogue,''
  in \emph{Proceedings of the 45th annual meeting of the association of
  computational linguistics}, 2007, pp. 496--503.

\bibitem{Furhat12}
S.~Al~Moubayed, J.~Beskow, G.~Skantze, and B.~Granstr{\"o}m, ``Furhat: A
  back-projected human-like robot head for multiparty human-machine
  interaction,'' in \emph{Cognitive Behavioural Systems}, A.~Esposito, A.~M.
  Esposito, A.~Vinciarelli, R.~Hoffmann, and V.~C. M{\"u}ller, Eds.\hskip 1em
  plus 0.5em minus 0.4em\relax Berlin, Heidelberg: Springer Berlin Heidelberg,
  2012, pp. 114--130.

\bibitem{bailenson2001equilibrium}
J.~N. Bailenson, J.~Blascovich, A.~C. Beall, and J.~M. Loomis, ``Equilibrium
  theory revisited: Mutual gaze and personal space in virtual environments,''
  \emph{Presence: Teleoperators \& Virtual Environments}, vol.~10, no.~6, pp.
  583--598, 2001.

\bibitem{reeves1996media}
B.~Reeves and C.~Nass, ``The media equation: How people treat computers,
  television, and new media like real people,'' \emph{Cambridge, UK}, vol.~10,
  p. 236605, 1996.

\end{thebibliography}
\end{document}